\documentclass[runningheads]{llncs}
\usepackage[T1]{fontenc}
\usepackage{graphicx}
\usepackage{color}

\usepackage{amsmath}
\usepackage{amsfonts}
\usepackage{wrapfig}
\usepackage{hyperref}

\begin{document}
\title{Fill in the Blanks: Accelerating Q-Learning with a Handful of Demonstrations in Sparse Reward Settings}
\titlerunning{Fill in the Blanks}
%
\author{Seyed Mahdi Basiri Azad\inst{1} \and
Joschka Boedecker \inst{1}}
\authorrunning{SM. Basiri Azad et al.}
%
\institute{
Faculty of Engineering University of Freiburg, Georges-Köhler-Allee 101 79110 Freiburg, Germany\\
\email{\{basiri,jboedeck\}@informatik.uni-freiburg.de}
}
%
\maketitle              

\begin{abstract}
Reinforcement learning (RL) in sparse-reward environments remains a significant challenge due to the lack of informative feedback. We propose a simple yet effective method that uses a small number of successful demonstrations to initialize the value function of an RL agent. By precomputing value estimates from offline demonstrations and using them as targets for early learning, our approach provides the agent with a useful prior over promising actions. The agent then refines these estimates through standard online interaction. This hybrid offline-to-online paradigm significantly reduces the exploration burden and improves sample efficiency in sparse-reward settings. Experiments on benchmark tasks demonstrate that our method accelerates convergence and outperforms standard baselines, even with minimal or suboptimal demonstration data.
\end{abstract}

\section{Introduction}
In many real-world applications of reinforcement learning (RL), agents must operate in environments where informative feedback is sparse or delayed. 
Sparse reward settings—where success is indicated by a binary signal, such as task completion or failure—pose a significant challenge for standard RL algorithms. 
In such environments, agents often receive no meaningful signal for extended periods, leading to slow and inefficient learning unless aided by strong inductive biases or extensive exploration.

A promising strategy to address this issue is to incorporate demonstration data. 
Demonstrations can guide agents toward reward-yielding regions of the state space, thereby reducing the need for blind exploration. 
However, existing approaches frequently assume access to a large set of expert trajectories or require detailed reward annotations—assumptions that are often unrealistic in practical deployments.

In this work, we introduce a simple yet effective method that jump-starts value learning using only a handful of successful demonstrations—potentially as few as one. 
Our approach makes no assumptions about demonstration optimality or access to dense reward signals. 
As long as the demonstrations achieve the task goal, they can offer valuable guidance. 
By estimating state value functions from demonstration data and using these estimates as targets for temporal difference (TD) updates, our method helps the agent focus its exploration early on. 
Once the agent begins interacting with the environment, it continues to refine its value estimates through standard TD learning in an online fashion.

Our method offers several key advantages:
\begin{enumerate}
    \item It is effective with minimal demonstration data—even a single successful trajectory can suffice.
    \item It accommodates sub-optimal demonstrations, as long as they achieve the task goal.
    \item It enables efficient learning in sparse-reward environments with only binary success/failure signals.
\end{enumerate}

We first motivate our approach through a regret analysis of $\epsilon$-greedy policies, highlighting how initial value estimates from demonstrations can reduce exploration cost. 
We then validate our method empirically across standard sparse-reward benchmarks, demonstrating substantial improvements in learning speed compared to vanilla Q-learning and common initialization baselines. 
Our results suggest that even minimal, imperfect demonstrations—if leveraged effectively—can dramatically improve sample efficiency in value-based RL.

\section{Related Works}

We review relevant literature on offline RL, fine-tuning from demonstrations, sparse reward settings, and learning from sub-optimal demonstrations.

\subsection{Offline RL from Demonstrations}

Offline RL methods~\cite{fujimoto2019off,kumar2020conservative,chen2021decision} learn policies solely from static data, leveraging rewards to potentially surpass the demonstrator. In contrast, imitation learning~\cite{bain1995framework,ross2011reduction,torabi2018behavioral} only mimics observed actions.

However, offline RL often assumes access to a well-defined reward function and large, diverse datasets—both costly in practice. Without environment interactions, these methods are also limited in their ability to improve beyond the demonstrations and tend to act conservatively to avoid out-of-distribution actions~\cite{kumar2020conservative}.

We instead use a few successful demonstrations to initialize Q-values before transitioning to online learning. This hybrid approach reduces the need for extensive offline data and enables continual policy improvement through interaction.

\subsection{Offline-to-Online RL (Warm-Starting / Fine-Tuning)}

Fine-tuning offline agents using online data aims to enhance robustness and generalization. However, most approaches rely on large offline datasets and suffer performance drops during the offline-to-online transition~\cite{ho2016generative,fu2018learningrobustrewards,nakamoto2023calql}, especially in sparse reward settings where mismatches in reward signals can destabilize value learning.

Our method avoids these pitfalls by using only a few successful demonstrations and explicitly supporting sparse rewards. By directly assigning value estimates to demonstrated states, we ensure a stable transition to online learning without reward mismatches or performance regressions.

\subsection{RL with Sparse Rewards}

Sparse reward settings simplify reward design but complicate learning due to infrequent feedback. Exploration methods using intrinsic rewards~\cite{pathak2017curiosity,tang2017exploration,lobel2023flipping,vieillard2020munchausen} encourage coverage but may not guide agents toward task completion.

Reward relabeling methods~\cite{andrychowicz2017hindsight,ho2016generative,dendorfer2020goalgan,azad2025srrewardtakingpathtraveled} generate synthetic goals, train discriminators between expert and non-expert data, or use expert visitation counts to densify feedback, but at the cost of added complexity and computation.

Instead, we reuse real demonstrations to increase informative transitions without altering the reward function or training auxiliary models. This preserves simplicity while facilitating goal-directed learning.

\subsection{Learning from Sub-optimal Demonstrations}

Since optimal demonstrations are hard to collect, many methods relax this requirement, using successful but sub-optimal trajectories~\cite{Rajeswaran2018learningcomplex,ball2023efficient}. These methods often schedule a gradual shift from imitation to reinforcement learning.

However, sparse rewards combined with limited demonstrations can lead to weak initial policies and hinder value propagation. Ranking-based approaches~\cite{brown2019extrapolating,lee2021pebble,rafailov2023direct} attempt to prioritize better trajectories but require many demonstrations and reliable ranking signals.

In contrast, our method requires only a few successful (not necessarily optimal) demonstrations. By precomputing value estimates for demonstrated states, we enable effective bootstrapping and allow online interactions to refine these initial estimates—without ranking or heavy data requirements.

\section{Background}
We now introduce the notation and review value-based reinforcement learning concepts. 

\subsection{Notation}
We consider settings where the environment is represented by a Markov Decision Process (MDP) and is defined as a tuple $\mathcal{M}=(\mathcal{S}, \mathcal{A}, \mathcal{T}, r, \gamma, \mu_0)$. $\mathcal{S}$ and $\mathcal{A}$ represent the state and action spaces respectively. 
$\mathcal{T}(s'|s, a)$ represents the state transition dynamics, $r(s, a)$ represents the reward function, $\gamma \in (0,1]$ is the discount factor and $\mu_0$ represents the starting state distribution. 
As in offline reinforcement learning setting, we only have access to a limited set of demonstrations of the form $\mathcal{D} = \{(s_0, a_0, s_{1}, a_{1},...s_T)^i\}_{i=0}^{N}$. 
In this paper, we are focusing on a limited setting where neither the transition dynamics $\mathcal{T}(s'|s,a)$ nor the dense reward function $r(s,a)$ are known.
The environment instead provides a sparse binary reward representing the success of failure of the agent in accomplishing a task.

\subsection{Reinforcement Learning}

In reinforcement learning (RL), an agent interacts with an environment modeled as a Markov Decision Process (MDP), aiming to learn a policy $\pi(a|s)$ that maximizes the expected cumulative discounted reward over time. 
The agent selects actions $a \in \mathcal{A}$ based on the current state $s \in \mathcal{S}$, receives a reward $r(s,a)$, and transitions to a new state $s' \sim \mathcal{T}(s'|s,a)$. 

The \emph{return} from a state $s_t$ is defined as the total discounted reward accumulated from time step $t$ onward:
\[
G_t = \sum_{k=0}^{\infty} \gamma^k r_{t+k},
\]
where $\gamma \in (0, 1]$ is a discount factor that prioritizes immediate rewards over distant ones.

To evaluate policies and guide decision-making, RL defines two primary value functions:

\begin{itemize}
    \item The \textbf{state-value function} $V^{\pi}(s)$ gives the expected return when starting in state $s$ and following policy $\pi$ thereafter:
    \[
    V^{\pi}(s) = \mathbb{E}_{\pi,\mathcal{T}}\left[ G_t \,\middle|\, s_t = s \right].
    \]
    
    \item The \textbf{action-value function} $Q^{\pi}(s,a)$ gives the expected return when taking action $a$ in state $s$ and following policy $\pi$ thereafter:
    \[
    Q^{\pi}(s,a) = \mathbb{E}_{\pi,\mathcal{T}}\left[ G_t \,\middle|\, s_t = s, a_t = a \right].
    \]
\end{itemize}

The optimal state-value and action-value functions, denoted by $V^*(s)$ and $Q^*(s,a)$ respectively, represent the maximum achievable expected return under any policy. The optimal policy $\pi^*$ then satisfies:
\[
\pi^*(s) = \arg\max_a Q^*(s,a).
\]

In value-based RL methods such as Q-learning, the goal is to learn an approximation of $Q^*(s,a)$, from which a near-optimal policy can be derived by acting greedily with respect to the learned values.

In sparse reward settings, where $r(s,a) = 0$ for most state-action pairs and only becomes non-zero upon success or failure, learning accurate value functions becomes particularly challenging due to the scarcity of meaningful feedback. Our proposed method addresses this challenge by leveraging a small number of successful demonstration trajectories to provide informative value estimates for visited states and actions, thereby significantly improving the sample efficiency of subsequent online learning.






\section{Methodology}

In environments with sparse rewards, value-based reinforcement learning algorithms often struggle to learn effective policies due to the difficulty of propagating reward signals through long trajectories. 
To address this, we propose a simple yet effective method to \textit{warm-start} Q-learning using a handful of successful demonstrations, even as few as one. 
Our approach initializes Q-values in states encountered in the demonstrations with informative estimates derived from observed returns. 

\subsection{Offline Value Estimation from Demonstrations}

We assume access to a small set of successful demonstration $\mathcal{D}$.
where each trajectory $\tau_i = (s_0, a_0, s_1, a_1, \dots, s_T)$ leads to a terminal success state and obtains a non-zero terminal reward (e.g., $r_T = 1$). 
The rewards at intermediate steps are assumed to be zero, consistent with a sparse reward setting.

Given this, we compute the return $G_t$ from each state $s_t$ in a demonstration as:
\[
G_t = \sum_{k=0}^{T - t} \gamma^k r_{t+k}
\]
Since rewards are zero until the terminal step and $r_T = 1$, the return simplifies to:
\[
G_t = \gamma^{T - t}
\]
We use this to estimate the state-value function $V(s_t)$ as:
\[
V(s_t) \leftarrow G_t
\]
for each state $s_t$ in the demonstration trajectories.

\subsection{Bootstrapping Q-values from Estimated V-values}

To warm-start Q-learning, we use these estimated $V(s)$ values to initialize the action-value function $Q(s,a)$ for all state-action pairs $(s,a)$ that appear in the demonstrations. 
Specifically, for every $(s_t, a_t) \in \tau_i$, we set:
\[
Q(s_t, a_t) \leftarrow V(s_t)
\]

This simple initialization provides a useful inductive bias: actions seen in the demonstrations are associated with values that reflect their ability to reach a successful outcome. 
Notably, this strategy does not require demonstrations to be optimal. 
In case of suboptimal, yet successful, demonstrations, the estimated value $V(s)$ represents a lower bound of the optimal state value $V^*(s)$. 
However, the closer the demonstration is to optimal, the closer our initialized $V(s)$ (and hence $Q(s,a)$) will be to their respective optimal counterparts, $V^*(s)$ and $Q^*(s,a)$.

\subsection{Online Fine-Tuning via Standard Q-learning}

After initializing Q-values for visited state-action pairs, the agent begins interacting with the environment. 
In the tabular case, the remaining Q-values (i.e., those not seen in the demonstrations) are initialized to zero. 
The agent then continues learning using standard tabular Q-learning:
\[
Q(s,a) \leftarrow Q(s,a) + \alpha \left( r + \gamma \max_{a'} Q(s', a') - Q(s,a) \right)
\]

Over time, the initialized Q-values help guide the agent's exploration toward rewarding trajectories, overcoming the cold-start problem that standard Q-learning faces in sparse reward settings.

\section{Analysis of the Regret Bound}
We can compare the expected regret of two $\epsilon$-greedy policies based on their initialized Q-values. 
Here, we assume a discrete state and action setting, deterministic environment dynamics, and access to optimal demonstrations. 
We use $s$ to represent a state from the state space $\mathcal{S}$ of size $|\mathcal{S}|$. 
Similarly, $a$ represents an action from the action space $\mathcal{A}$ of size $|\mathcal{A}|$.
A state $s_{demo}$ is part of the demonstration states $\mathcal{S}_{demo}$, and $a^*$ represents the optimal action at each state.

Let's define regret as the following:
\begin{equation} \label{eq:cumulative-regret}
    \text{Regret} = \frac{1}{|\mathcal{S}|}\sum_{s} \sum_{a} \pi(a|s)\mathbb{I}(r(s,a^*) - r(s,a))
\end{equation}

$\mathbb{I}$ is an indicator function that is zero if the optimal action is taken in state $s$ and one otherwise. This is a simplified way to measure regret in discrete action spaces, especially when we care more about suboptimal action frequency than the magnitude of suboptimality.

\subsection{Case 1: Q values initialized with zeros (No Demonstration)}
$\epsilon$-greedy policy reverts to a uniform random policy when all Q-values are zero:
\begin{equation*}
    \pi(a|s) = \frac{1}{|\mathcal{A}|}
\end{equation*}
Regret in any state $s \in \mathcal{S}$ will be:
\begin{equation*}
    \begin{split}
        \text{Regret}(s) &= \sum_a\pi(a|s)\mathbb{I}(r(s,a^*) - r(s,a)) = \frac{1}{|\mathcal{A}|} \sum_a\mathbb{I}(r(s,a^*) - r(s,a)) = \frac{|\mathcal{A}|-1}{|\mathcal{A}|}\\
    \end{split}
\end{equation*}
All actions create regret except $a=a^*$.
So the regret over all states will be:
\begin{equation*}
    \begin{split}
        \text{Regret} &= \frac{1}{|\mathcal{S}|}\sum_s \text{Regret(s)} = \frac{1}{|\mathcal{S}|}|\mathcal{S}|( \frac{|\mathcal{A}|-1}{|\mathcal{A}|}) = ( \frac{|\mathcal{A}|-1}{|\mathcal{A}|})
    \end{split}
\end{equation*}

\subsection{Case 2: Q values partially initialized (With Demonstration)}
In this scenario, we have to split the states into the ones that appear in the demonstrations, $s_{demo} \in \mathcal{S}_{Demo}$, and those that do not, $s \notin \mathcal{S}_{Demo}$.
For $s_{demo} \in \mathcal{S}_{Demo}$ the policy will act like an $\epsilon$-greedy policy:

\begin{equation*}
    \pi(a|s_{demo}) =     
    \begin{cases}
    (1-\epsilon) + \frac{\epsilon}{|\mathcal{A}|} & if a = a^*\\
    \frac{\epsilon}{|\mathcal{A}|} & otherwise
    \end{cases}
\end{equation*}
For non-demo states, the policy will act like a uniform random policy as before:

\begin{equation*}
    \pi(a|s) = \frac{1}{|\mathcal{A}|}
\end{equation*}
So regret for a state in the demonstrations will be:

\begin{equation*}
    \begin{split}
        \text{Regret}(s_{demo}) &= \sum_a\pi(a|s_{demo})\mathbb{I}(r(s_{demo},a^*) - r(s_{demo},a)) = \epsilon - \frac{\epsilon}{|\mathcal{A}|}
    \end{split}
\end{equation*}
This is because $(1-\epsilon) + \frac{\epsilon}{|\mathcal{A}|}$ of the time the optimal action $a^*$ will be taken, which results in no regret.
The regret over all demonstration states will then be:
\begin{equation*}
    \begin{split}
        \text{Regret}_{Demo} &= \frac{1}{|\mathcal{S}|}\sum_{s_{demo} \in \mathcal{S}_{Demo}}\text{Regret}(s_{demo}) = \frac{1}{|\mathcal{S}|}|\mathcal{S}_{Demo|}(\epsilon - \frac{\epsilon}{|\mathcal{A}|})
    \end{split}
\end{equation*}
The regret over all non-demo states will be based on the random uniform policy:

\begin{equation*}
    \begin{split}
        \text{Regret}_{NonDemo} &= \frac{1}{|\mathcal{S}|}\sum_{s \notin \mathcal{S}_{Demo}}\text{Regret}(s) = \frac{1}{|\mathcal{S}|}(|\mathcal{S}|-|\mathcal{S}_{Demo}|)(\frac{|\mathcal{A}|-1}{|\mathcal{A}|})
    \end{split}
\end{equation*}
So the overall regret for this case will be:
\begin{equation*}
    \begin{split}
        \text{Regret} &= \frac{1}{|\mathcal{S}|}(\text{Regret}_{Demo} + \text{Regret}_{NonDemo}) \\
        &=\frac{1}{|\mathcal{S}|}(|\mathcal{S}_{Demo|}(\epsilon - \frac{\epsilon}{|\mathcal{A}|}) + (|\mathcal{S}|-|\mathcal{S}_{Demo}|)(\frac{|\mathcal{A}|-1}{|\mathcal{A}|}))
    \end{split}
\end{equation*}

 \subsection{Grid-World Example}
 
\begin{wrapfigure}{r}{0.25\textwidth}
    \centering
    \vspace{0pt}
    \includegraphics[width=0.25\textwidth]{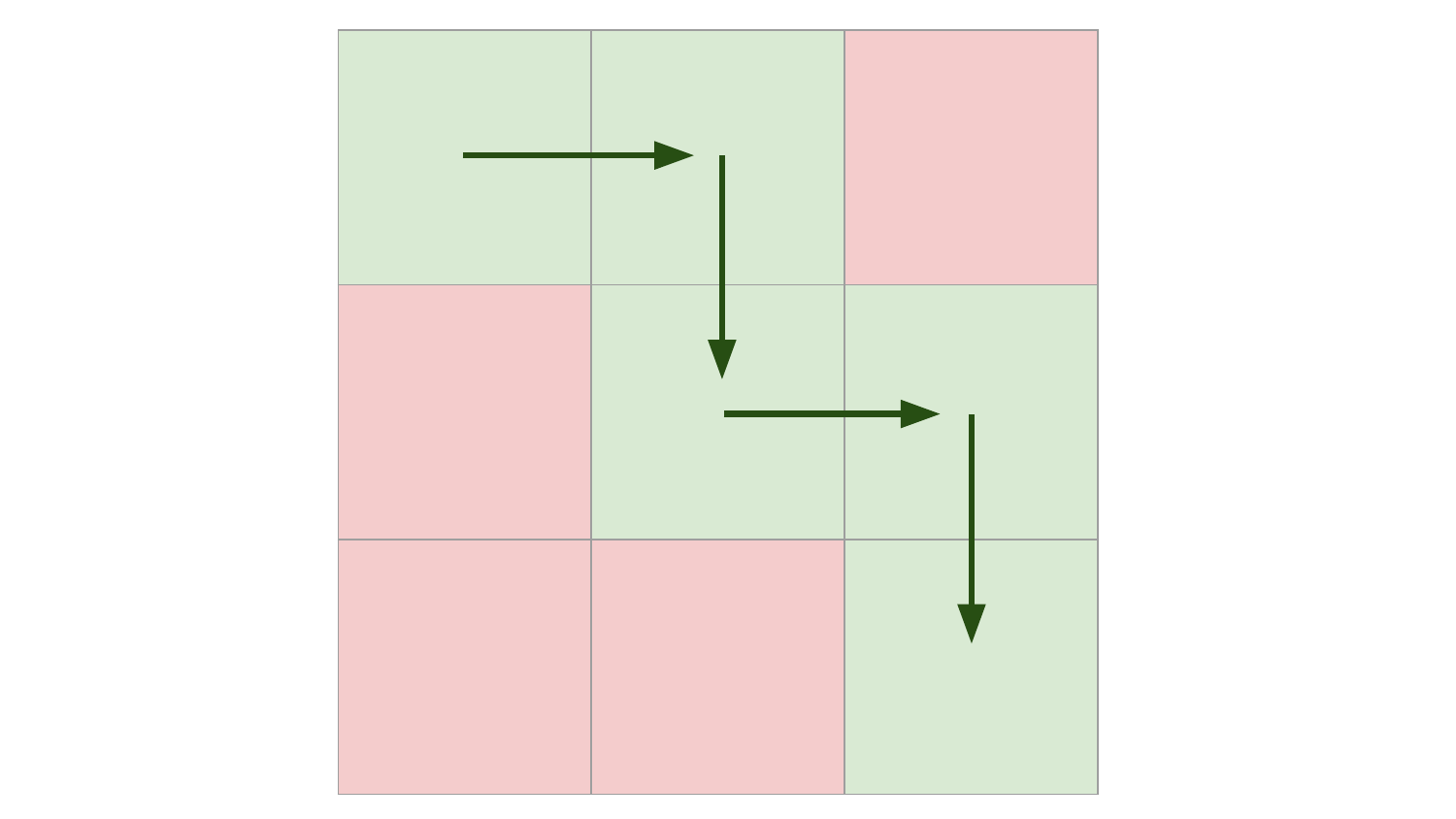}
    \caption{3x3 grid with one demonstration}
    \label{fig:3x3grid}
    \vspace{-0pt}
\end{wrapfigure}

To better understand the impact of partial Q initialization, let's consider a simple 3x3 grid-world in Figure~\ref{fig:3x3grid}. 
We assume that the states wrap around, so there are 4 actions available in each state, and the $\epsilon=0.1$.

Without demonstrations:
\begin{equation*}
    \text{Regret} = ( \frac{|\mathcal{A}|-1}{|\mathcal{A}|}) = (\frac{4-1}{4}) = \textbf{0.75}
\end{equation*}

With the demonstration:
\begin{equation*}
    \begin{split}
        \text{Regret} &= \frac{1}{|\mathcal{S}|}(|\mathcal{S}_{Demo|}(\epsilon - \frac{\epsilon}{|\mathcal{A}|}) + (|\mathcal{S}|-|\mathcal{S}_{Demo}|)(\frac{|\mathcal{A}|-1}{|\mathcal{A}|}))\\
        &=\frac{1}{9} (5(0.1-\frac{0.1}{4}) + (9-5)(\frac{4-1}{4})) = \textbf{0.375}
    \end{split}
\end{equation*}

In this example, the expected regret was reduced by almost half using the demonstrations. One of the most important factors affecting the expected regret as presented by Eq.~\ref{eq:cumulative-regret} is the number of states visited in the demonstrations. Eq.~\ref{eq:cumulative-regret} suggests a sharp decrease in the advantage gained by partial initialization of the Q-values as the ratio of demonstration states to state space decreases. 

Interestingly, our empirical findings suggest a continuous benefit from partially initializing the Q-values, even from a single demonstration, particularly as environments grow larger and more complex. 
The discrepancy arises because Eq.~\ref{eq:cumulative-regret} does not account for the shift in the on-policy state distribution that occurs as the policy improves through partial Q initialization. 
As such, Eq.~\ref{eq:cumulative-regret} should be interpreted as an \textbf{upper bound on the expected regret}, rather than a tight characterization of the empirical performance.
The true expected regret can be formulated as:
\begin{equation} \label{eq:expected-regret}
    \text{Expected-Regret} = \sum_{s} d^{\pi}(s) \sum_{a} \pi(a|s)\mathbb{I}(r(s,a^*) - r(s,a))
\end{equation}

with $d^{\pi}$ representing the on-policy state distribution implied by the behavior of policy $\pi$. Partial initialization of the Q-values from successful demonstrations increases the likelihood of the resulting policy staying close to the demonstrations where the regret is minimum. We discuss our empirical results further in section \ref{sec:tab_q}.

\section{Beyond Tabular Q-Learning}\label{sec:beyondTabQL}

In continuous state and action spaces, function approximation is essential for representing value functions and policies. 
Neural networks, in particular, are powerful approximators but introduce a range of new challenges when extending tabular Q-learning methods to continuous domains.

A key issue is \textbf{extrapolation error}, which arises when neural networks encounter out-of-distribution inputs. 
When assigning Q-values to state-action pairs along demonstration trajectories, we treat the problem as a regression task.
However, because the training data covers only a narrow subset of the space—primarily successful trajectories from a small number of demonstrations—Q-values for unseen samples may be significantly overestimated. 
This can mislead the policy to explore regions far from the demonstrations, rather than nearby areas with higher expected returns.

Another challenge is \textbf{catastrophic forgetting}, where the network gradually overwrites useful Q-values obtained from demonstrations as new transitions are collected during online learning. 
This is exacerbated in sparse reward settings, where most online samples contain no reward, pushing regression targets toward zero and causing the network to forget valuable prior knowledge.

Unlike supervised learning on static datasets, reinforcement learning involves continuously updating value functions toward moving targets that often increase over time. 
This dynamic gives rise to \textbf{primacy bias}—a tendency for neural networks to overfit early experiences and underutilize later, more informative data~\cite{Nikishin2022ThePB}.

To mitigate overestimation, we reformulate Q-value regression as a classification task. 
Prior work has shown that categorical distributions can stabilize value learning and better track moving targets~\cite{bellemare2017adistributional,farbebrother2024stopreg}. 
Following~\cite{farbebrother2024stopreg}, we discretize the Q-value range into fixed bins, treating value estimation as a categorical prediction task. 
While this approach may require prior knowledge of value ranges in dense reward settings, it is well suited to sparse settings where the maximum Q-value is known and bounded. 
This formulation not only reduces extrapolation error but also improves stability during training as the policy evolves.

Although prior work has suggested partially resetting networks to reduce primacy bias~\cite{Nikishin2022ThePB}, we find this unnecessary in our setup. 
The bounded value range inherent in sparse reward settings, combined with the categorical value formulation, provides sufficient stability and adaptability to mitigate primacy bias in practice.

To address forgetting, we maintain separate replay buffers for offline and online data. 
During training, we use two updates: one batch from online transitions and one from demonstration data, with sparse cumulative rewards replacing instantaneous ones for consistency.

\subsection{On the Importance of the Discount Factor $\gamma$}
The discount factor $\gamma$ is a critical parameter in our method, as it directly influences the value estimates derived from demonstration data. 
When estimating the value of a state based on future rewards in a demonstration, $\gamma$ determines how much value is propagated backward to earlier states.
In tasks that require many steps to reach a goal, a higher $\gamma$ is necessary to preserve meaningful value estimates for the initial states. 
For instance, consider a task that typically takes 100 steps to complete. 
With $\gamma = 0.99$, the estimated value of the initial state would be $\gamma^{100} = 0.366$, whereas with $\gamma = 0.95$, this value drops drastically to $\gamma^{100} = 2 \times 10^{-5}$. 

Such low values can become indistinguishable from the noisy outputs of an untrained value function approximator, effectively masking the only available learning signal in a sparse reward setting. 
As a result, the agent may struggle to benefit from the demonstration guidance. 
This highlights the importance of selecting a sufficiently high $\gamma$ to ensure that value signals from sparse rewards can propagate far enough to meaningfully influence learning from demonstrations.

\section{Experiments}
The goal of our experiments is to gain a better understanding of the advantages of partial Q-initialization and show how our choice of techniques enables the use of partial Q initialization to continuous state and action settings via function approximation. We then evaluate our method using three seeds on popular simulated environments~\cite{towers2024gymnasium}.


\subsection{Tabular Q-Learning} \label{sec:tab_q}
We use the \textit{FrozenLake} environment to empirically estimate the expected regret. This environment uses a discrete state and action space and provides only a sparse reward when reaching a successful state. 
We modify the complexity of the tasks by creating obstacles and extending the size of the maze. 

\begin{figure}[tb]
    \centering
    \begin{tabular}{llll}
        \includegraphics[width=0.14\linewidth]{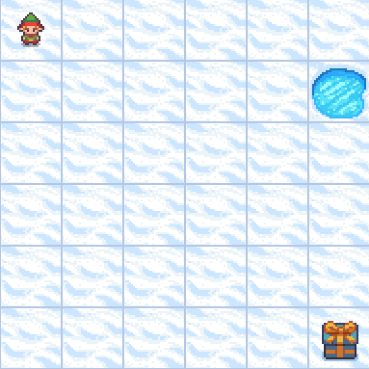} & 
        \includegraphics[width=0.15\linewidth, trim={60 0 60 0},clip]{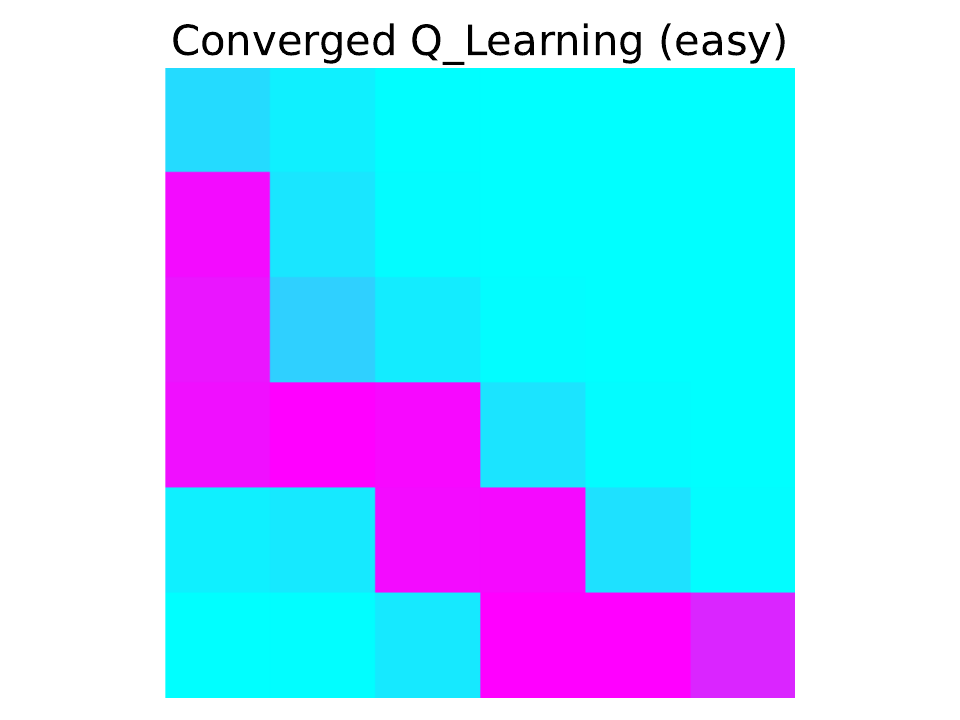} &
        \includegraphics[width=0.15\linewidth, trim={60 0 60 0},clip]{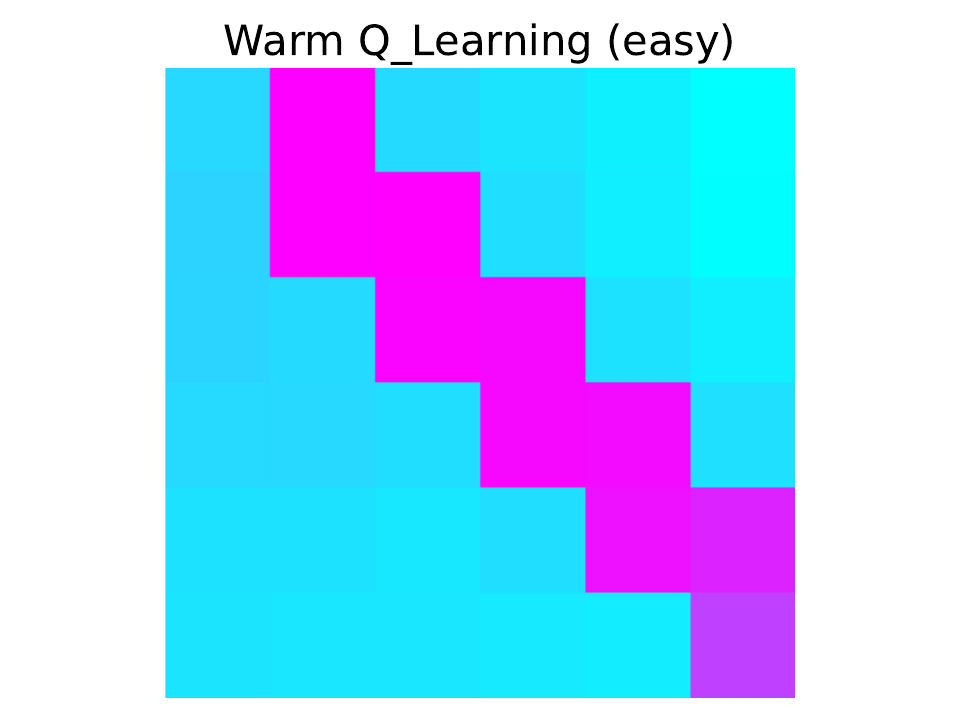} &
        \includegraphics[width=0.15\linewidth, trim={60 0 60 0},clip]{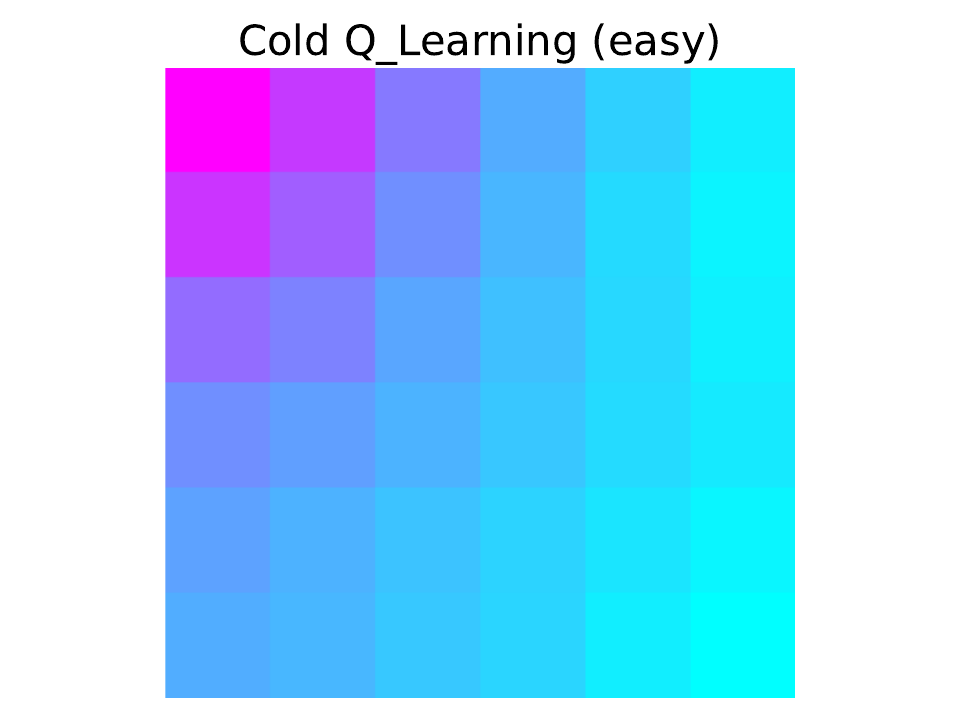} \\

        \includegraphics[width=0.14\linewidth]{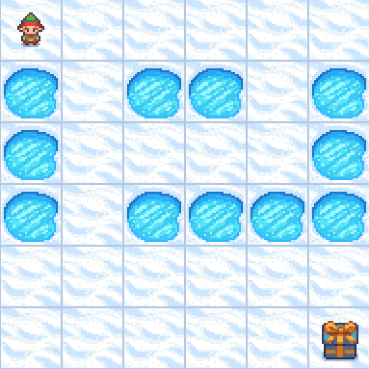} & 
        \includegraphics[width=0.15\linewidth, trim={60 0 60 0},clip]{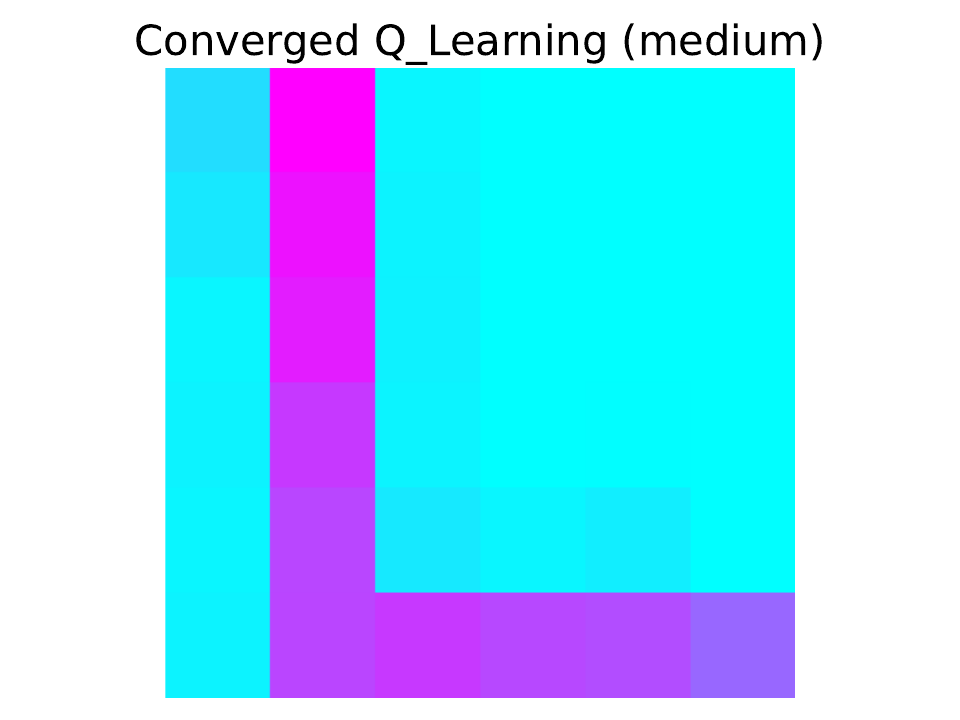} &
        \includegraphics[width=0.15\linewidth, trim={60 0 60 0},clip]{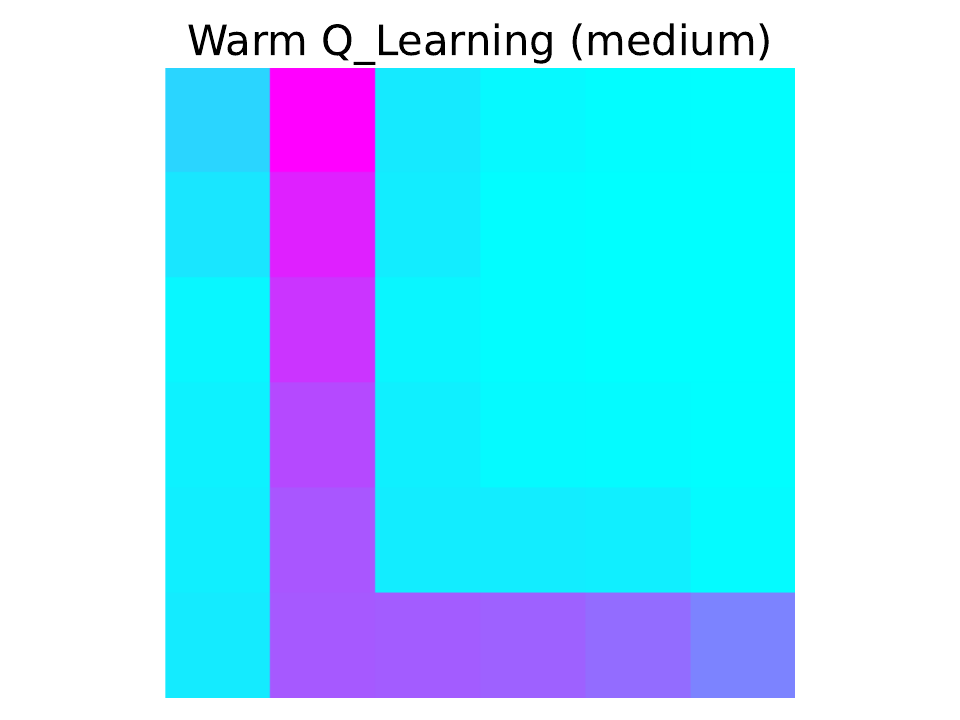} &
        \includegraphics[width=0.15\linewidth, trim={60 0 60 0},clip]{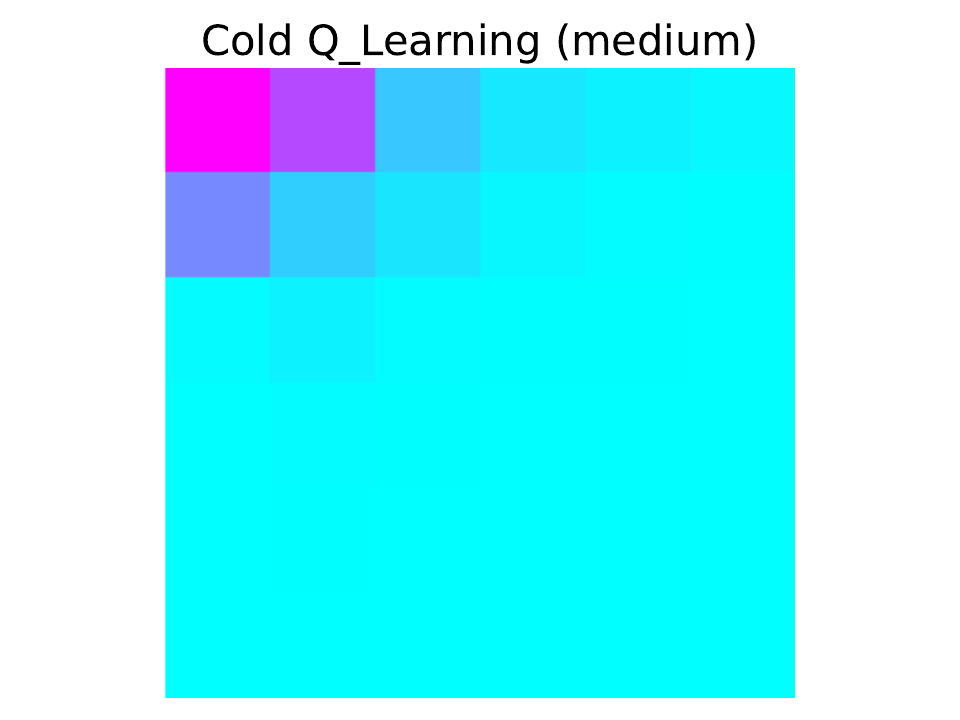} \\

        \includegraphics[width=0.14\linewidth]{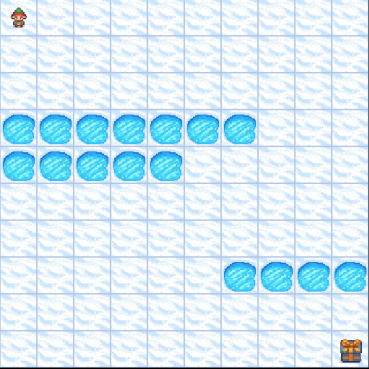} &
        \includegraphics[width=0.15\linewidth, trim={60 0 60 0},clip]{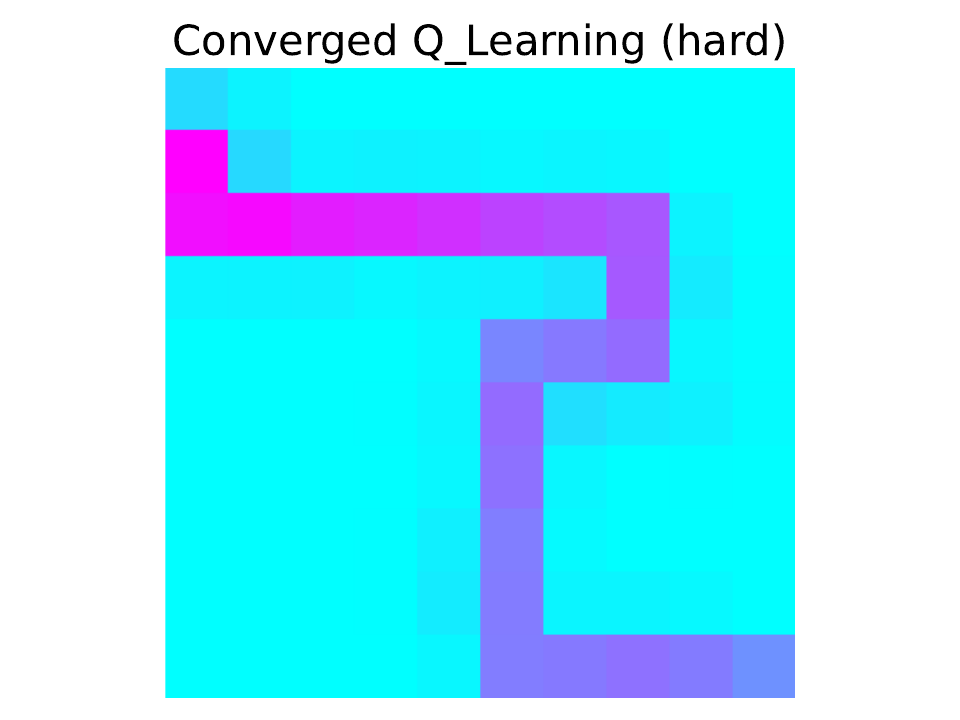} &
        \includegraphics[width=0.15\linewidth, trim={60 0 60 0},clip]{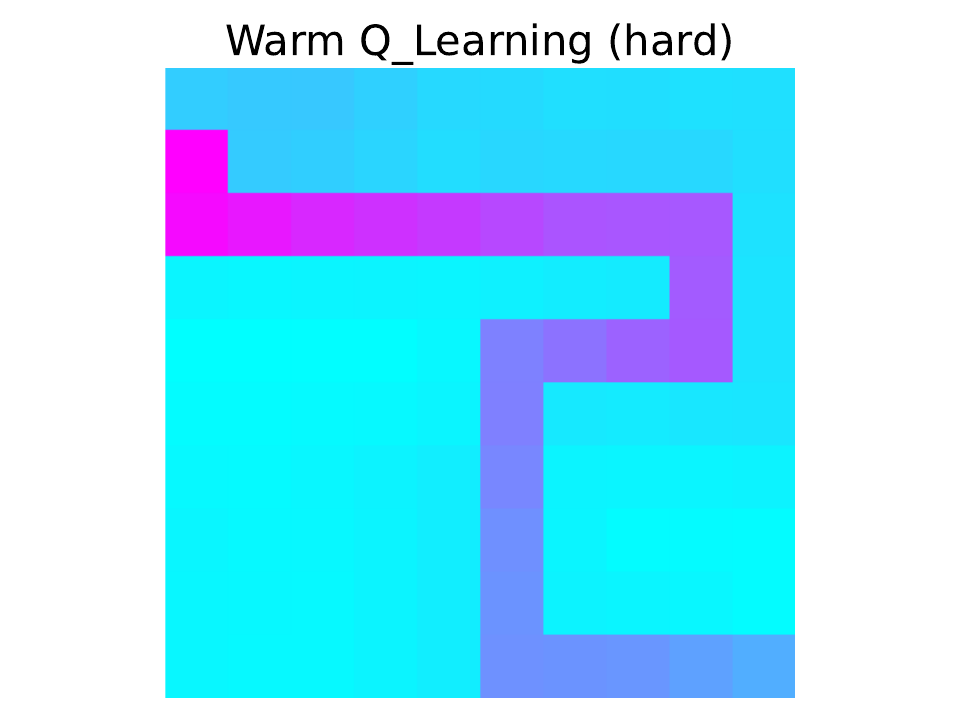} &
        \includegraphics[width=0.15\linewidth, trim={60 0 60 0},clip]{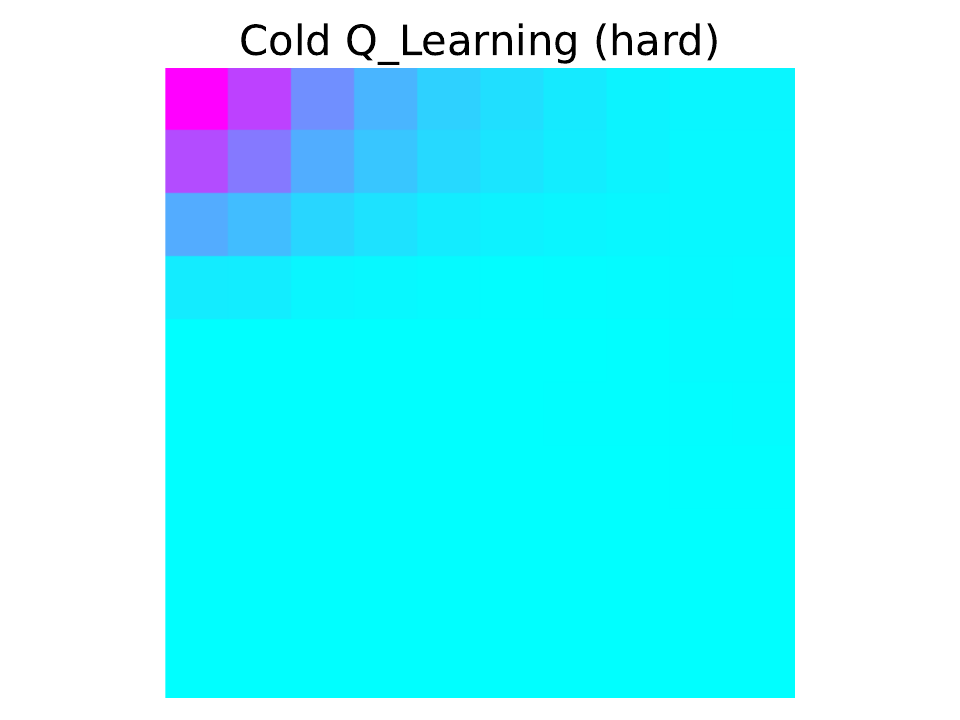} \\

    \end{tabular}
        
    \caption{On-policy state visitation for Easy (top), Medium (middle), and Hard (bottom) versions of the environment. The left column shows the environments with ice holes represented in blue. The following columns show the on-policy state visitation probability implied by the policies from Converged Q-Learning (second column), Demo-initialized Q-Learning (third column), and zero-initialized Q-Learning (fourth column.}
    \label{fig:on-policy-dist}
\end{figure}

Figure~\ref{fig:on-policy-dist} shows the on-policy state distribution for policies derived from converged, partially initialized (demo initialized), and uninitialized Q-tables. 
The converged Q-Learning agent is trained using 100,000 episodes with a maximum number of steps per episode set to 100. We use a single demonstration to partially initialize the Q-table. The resulting $\epsilon$-greedy policies with $\epsilon=0.2$ are then used for 1000 rollouts, and the number of visits for every state is recorded and normalized to calculate the on-policy state distribution of each policy. 
Figure~\ref{fig:on-policy-dist} shows that agents fail to explore the environment efficiently in the beginning, a condition that is exacerbated with the growing number of obstacles and the size of the state space. Initializing the Q-values, even with a single demonstration, guides the exploration of the agent in the regions near the demonstration. This is especially helpful in settings with sparse rewards, as random exploration is unlikely to result in successful completion of the task.  

Table~\ref{tab:regrets} shows the expected regret for $\epsilon$-greedy policies based on Q-tables with different initialization. The expected regret is calculated based on Eq.~\ref{eq:expected-regret} using the optimal action from the converged Q-table. The expected regrets are calculated once assuming a uniform on-policy state distribution and once using the actual distribution of the policies(Figure~\ref{fig:on-policy-dist}). Once the actual state distribution is taken into account, the expected regret decreases significantly for policies based on demo-initialized Q even as the environment's state space and difficulty increase.

\begin{table}
\caption{Regret for different $\epsilon$-greedy policies based on uniform or true state distribution.}
\label{tab:regrets}
\centering
\begin{tabular}{|l|c|c|c|c|}
\hline
Difficulty & Warm Q (Uniform) & Warm Q (True) & Cold Q (Uniform) & Cold (True) \\
\hline
 Easy & 0.70 & \textbf{0.59} & 0.75 & 0.76 \\
 Medium & 0.58 & \textbf{0.26} & 0.75 & 0.75 \\
 Hard & 0.62 & \textbf{0.38} & 0.75 & 0.74 \\
\hline
\end{tabular}
\end{table}

\subsection{Continuous State and Action Spaces}
As discussed in Section~\ref{sec:beyondTabQL}, extending our approach to continuous state and action spaces presents new challenges when using function approximations. 
In this section, we empirically validate the effectiveness of our design choices in addressing these challenges and demonstrate that our method generalizes beyond the tabular setting.

\begin{figure}[tb]
    \centering
    \includegraphics[width=0.49\linewidth]{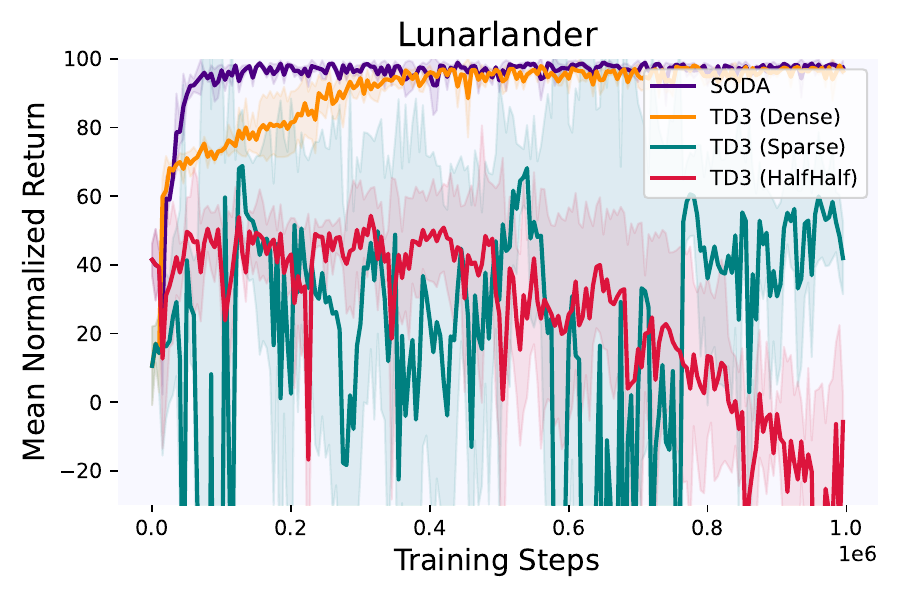}
    \includegraphics[width=0.49\linewidth]{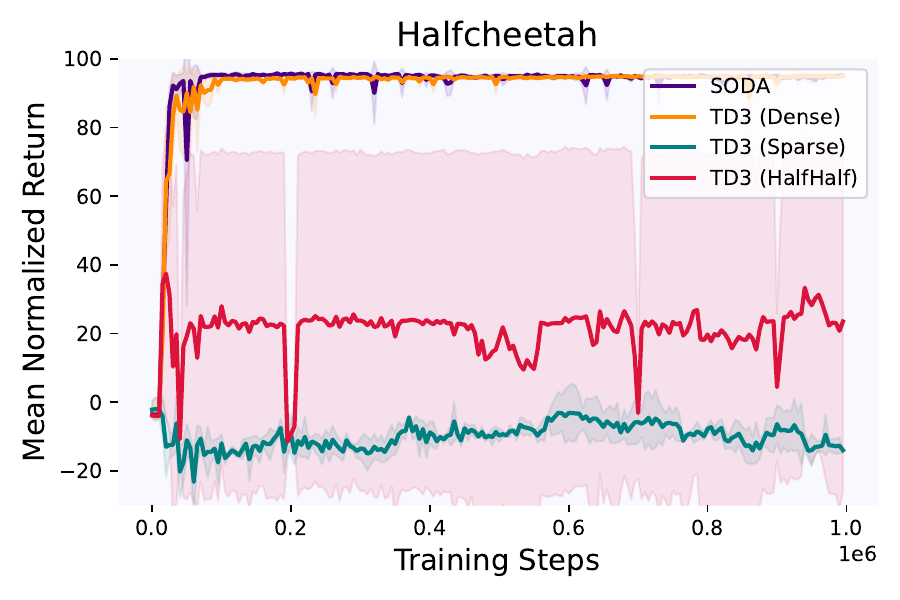}
    \caption{SODA can accelerate the learning, especially in sparse reward settings.}
    \label{fig:validations}
\end{figure}

Figure~\ref{fig:validations} compares our method—\textbf{S}parse \textbf{O}ffline-Online \textbf{D}emonstration-based \textbf{A}cceleration (SODA)—against TD3 on the LunarLander and HalfCheetah environments.
SODA is trained using only 10 expert demonstrations. We include TD3 with dense rewards as an optimistic upper bound, noting that our primary focus is on sparse reward scenarios.
While TD3 performs well with dense rewards, its performance degrades significantly in the sparse reward setting, failing to solve the tasks. In contrast, SODA demonstrates substantially faster learning and convergence under sparse rewards, highlighting the benefit of initializing value estimates from demonstrations.
We also compare against prior approaches~\cite{ball2023efficient,Rajeswaran2018learningcomplex} that incorporate demonstrations by composing each training batch with 50\% offline and 50\% online transitions. As shown in Figure~\ref{fig:validations}, this hybrid batching strategy offers limited benefit in sparse reward settings, since most transitions—including those from demonstrations—lack meaningful reward signals for value learning.
These results underscore the importance of directly assigning value estimates to demonstrated states, rather than simply mixing offline and online data.

\subsection{Ablation: Number of Demonstrations}
\begin{figure}[tb]
    \centering
    \includegraphics[width=0.49\linewidth]{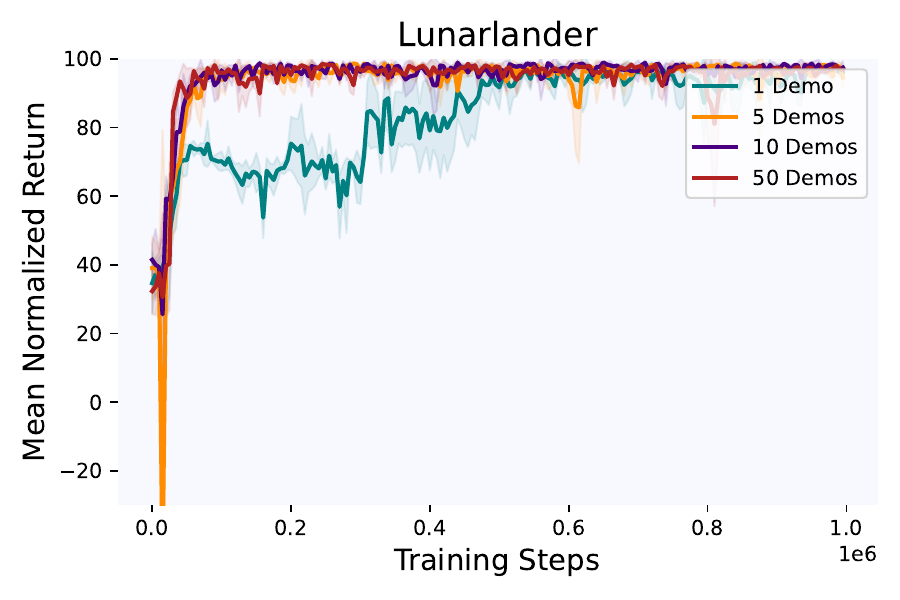}
    \includegraphics[width=0.49\linewidth]{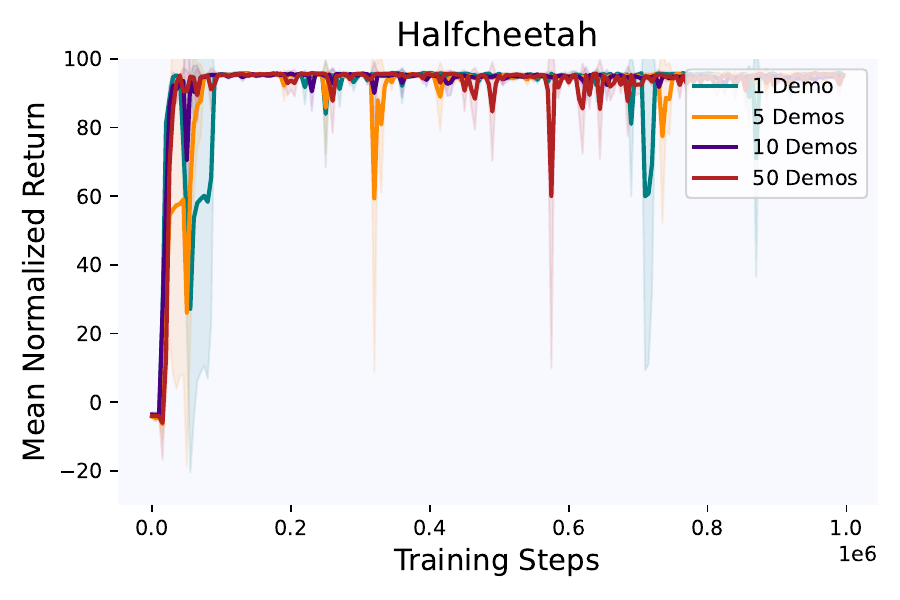}
    \caption{SODA can learn with as few as 1 demonstration.}
    \label{fig:ablation-demos}
\end{figure}

As implied by Eq.~\ref{eq:expected-regret}, increasing the number of demonstrations should reduce regret by covering a larger portion of the state space. 
However, even a single successful demonstration can shift the on-policy state distribution and effectively guide the actor toward regions that yield sparse rewards.
We validate this hypothesis by training SODA with varying numbers of demonstrations: [1, 5, 10, 50]. 
As shown in Figure~\ref{fig:ablation-demos}, even a single demonstration enables SODA to solve the task under sparse rewards. 
As expected, increasing the number of demonstrations further accelerates convergence, but the strong performance with minimal data highlights the efficiency of our approach.

\subsection{Ablation: Demonstration Quality}
\begin{figure}[tb]
    \centering
    \includegraphics[width=0.49\linewidth]{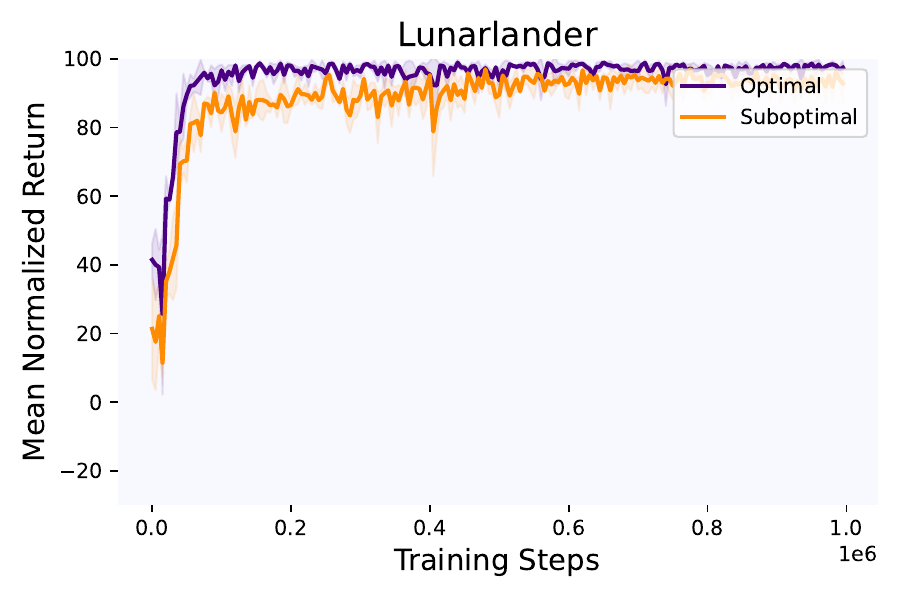}
    \includegraphics[width=0.49\linewidth]{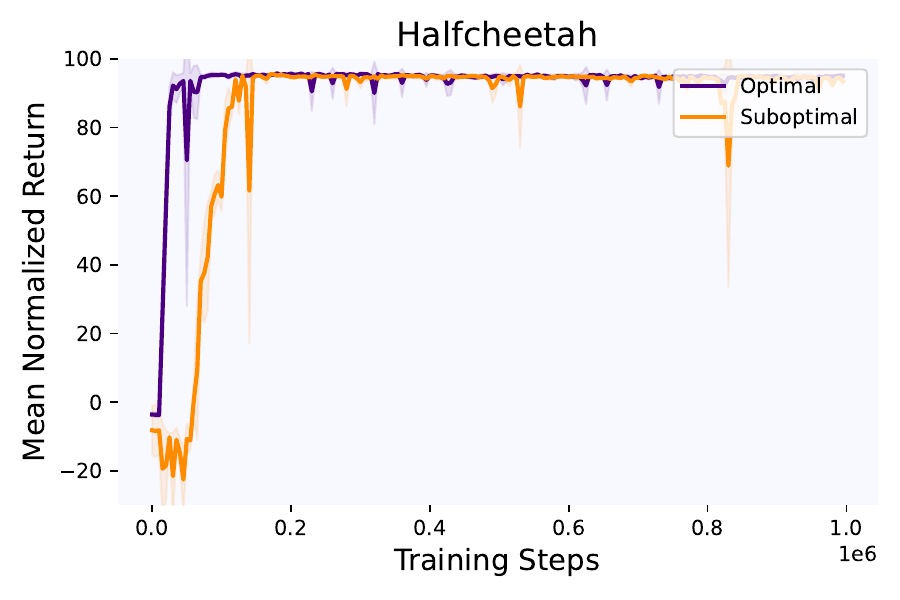}
    \caption{SODA can learn to solve the environments using sub-optimal demonstrations.}
    \label{fig:ablation-optimality}
\end{figure}

Even sub-optimal but successful demonstrations can provide useful initial estimates for value functions—especially relevant in settings where demonstrations come from human teleoperation or motion planning. 
Figure~\ref{fig:ablation-optimality} compares SODA trained with 10 optimal versus sub-optimal demonstrations.
Optimal demonstrations are generated using a fully trained TD3 agent with access to dense rewards, while sub-optimal ones are collected from a TD3 agent stopped early, once it reliably but inefficiently completes the task.
While sub-optimal demonstrations may slow convergence (e.g., in HalfCheetah) or slightly affect final performance (e.g., in LunarLander), they still significantly improve exploration and enable successful learning in sparse reward settings. 
This highlights the robustness of our method to demonstration quality.

\section{Conclusion}
We presented a method to accelerate reinforcement learning in sparse reward settings by explicitly estimating value functions for states visited in a small set of demonstrations and using these estimates as targets for temporal difference updates. 
Through regret analysis, we motivated the effectiveness of this approach in the tabular Q-learning setting and outlined a set of design choices that enable its extension to continuous state and action spaces via function approximation.

Our empirical results demonstrate that these choices lead to significant improvements in learning speed and stability. 
Ablation studies further confirm the robustness of our method to variations in the number and quality of demonstrations. 
While our empirical results are limited to relatively simple environments, they establish a strong foundation for the method. Future work will focus on evaluating the approach in more complex, high-dimensional robotics tasks.

\begin{credits}
\subsubsection{\ackname}
This work was funded by Carl Zeiss Foundation with the ReScaLe
project.
The authors have no competing interests to declare that are
relevant to the content of this article.
\end{credits}
%
%
%
\bibliographystyle{splncs04}
\bibliography{mybibliography}
%




\end{document}